\documentclass[letterpaper, 10 pt, journal, twoside]{IEEEtran}  %
\usepackage{generic}
\usepackage{cite}
\usepackage{amsmath,amssymb,amsfonts}
\usepackage{algorithmic}
\usepackage{graphicx}
\usepackage{textcomp}

\usepackage{amsmath} % assumes amsmath package installed

\usepackage{url}
\usepackage[table,x11names,dvipsnames,table]{xcolor}
\definecolor{lightgray}{gray}{0.95}
\usepackage{lipsum}
\usepackage{array} % 导言区
\usepackage{color}
\usepackage[ruled,vlined]{algorithm2e}
\usepackage{diagbox} 
\usepackage{tabu}
\usepackage{arydshln} 
\usepackage{makecell} % 导言区引入

\usepackage{float,tablefootnote,threeparttable,wrapfig}
\usepackage{booktabs}
\usepackage{bm}
\usepackage{subcaption}

\usepackage{framed,multirow}

\begin{document}
\title{Point-Wise Geometry-Aware Transformer for Partial-to-Full Point Cloud Registration in Computer-Assisted Surgery}
\author{
Siyu Zhou, and Zhongliang Jiang
\thanks{
The Chair for Computer Aided Medical Procedures, Technical University of Munich, Munich, Germany. 
The authors are now affiliated with The University of Hong Kong, Hong Kong SAR, China
}
\thanks{
This work involved human subjects in its research. Approval of all ethical and experimental procedures and protocols was granted by Institutional Review Board, No. 2022-87-S-KK, Declaration of Helsinki. 
}
}

\maketitle

\begin{abstract}
Partial-to-full registration remains challenging due to varying overlap ratios, fluctuating point densities, and the presence of noise. While transformers have shown strong potential for point cloud processing, prior methods typically confine them to global context aggregation, overlooking fine-grained local geometry crucial for accurate correspondence. We propose \emph{GAPR-Net}, a learning-based point cloud registration framework with a coarse-to-fine architecture that combines convolution and transformer modules, in which local and global information is fused between the partial and full point clouds using a cross-attention mechanism. To achieve this, a transformation-invariant point-wise geometric feature representation is proposed, which can robustly capture relative geometric features for individual points with respect to their neighboring points. To evaluate the effectiveness of the proposed approach, experiments are conducted on four geometrically distinct bones, including the tibia, femur, pelvis, and thoracic cartilage. The overall registration recall reaches 94.2\%, the method results in a low RMSE of 1.992 mm and R² values of 0.908 and 0.974 for rotation and translation, respectively.  The results demonstrate that the proposed method effectively addresses the partial-to-full point cloud registration problem. The proposed method enables highly accurate 3D point cloud registration using partial observation, providing a critical foundation for precise surgical navigation and robotic interventions in computer-assisted surgery. The code will be accessed after the double-blind review process.
% The code can be found on this website: \url{https://github.com/howling1/PGA-for-Partial2Full-Reg}.
\end{abstract}

% \def\abstractname{Note to Practitioners}
% \begin{abstract}
% The proposed method provides a learning-based solution for robust point cloud registration under low-overlap and partial-to-full settings by combining local geometric features with global context in a coarse-to-fine framework. Its primary application is in computer-assisted medical systems, where accurate registration is critical for surgical navigation and intervention, improving reliability under sparse and noisy observations. However, the current approach is validated mainly on pre-segmented bone structures. Future work should address real-time performance and extend the method to soft tissues and multimodal sensing. Beyond medical applications, the framework can also be applied to scenarios with partial observations, such as scene or home reconstruction, offering a step toward more scalable and dependable 3D perception.
% \end{abstract}

\begin{IEEEkeywords}
Partial-to-full registration; Computer-assisted surgery; ultrasound CT registration;
\end{IEEEkeywords}

\section{Introduction}
\par
In modern computer-assisted surgery (CAS), accurate registration is essential for transferring the planned surgical trajectory from pre-operative imaging modalities, such as CT or MRI, to intra-operative data, e.g., ultrasound scans. This alignment ensures that the surgical plan corresponds accurately to the patient-specific layout during the procedure~\cite{zhang2025deepbhmr,zhou2025directed}. However, since intra-operative data often provides an incomplete view compared to comprehensive pre-operative images, the task becomes a complex partial-to-full registration problem. This challenge involves aligning two datasets with limited overlap, requiring advanced techniques to achieve accurate and reliable results in such high-stakes scenarios~\cite{zhang2024ghmm, chen2023generalized, lu2023preoperative}.

%%%%%%%%%%%%%%%%%%%%%%%%%
\begin{figure}[ht!]
\centering
\includegraphics[width=0.48\textwidth]{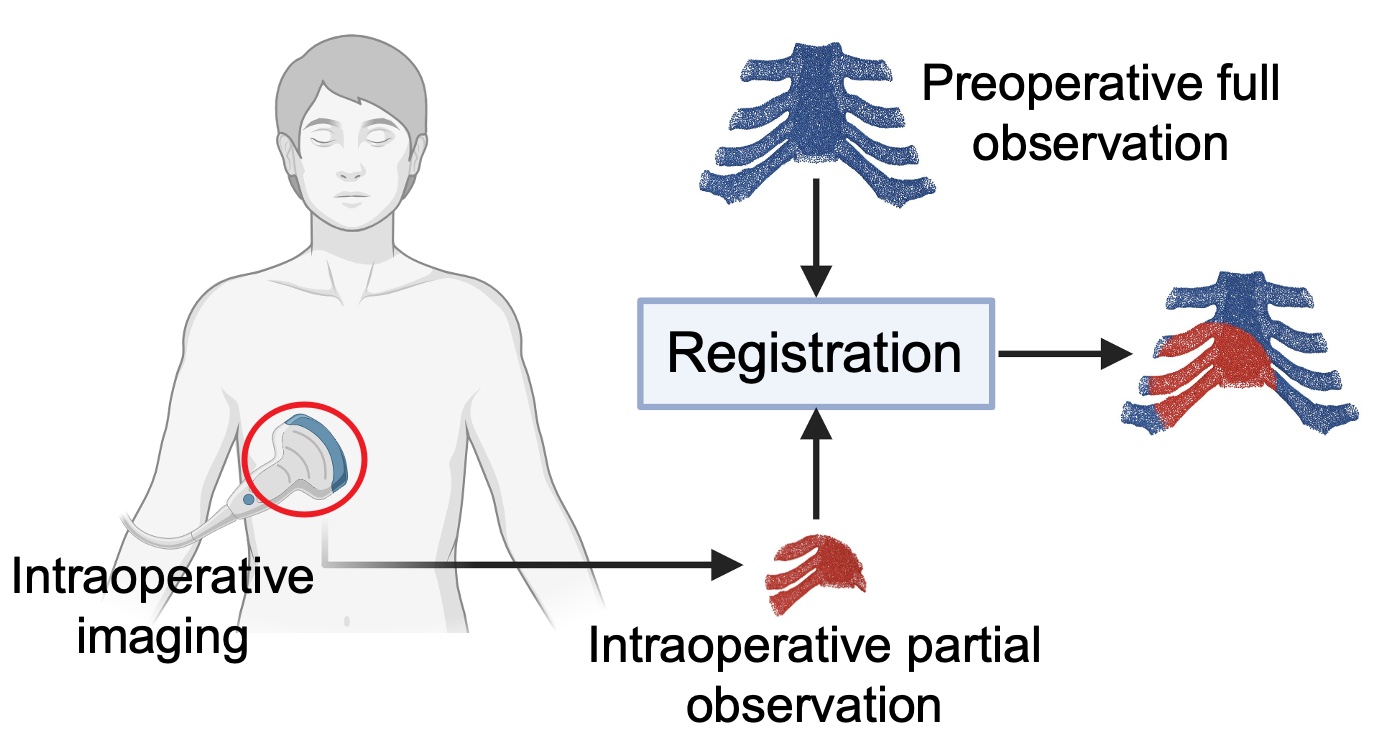}
\caption{An illustration of the thoracic surgical application based on the thoracic cartilage partial-to-full point cloud registration.
}
\label{Fig_teaser}
\end{figure}
%%%%%%%%%%%%%%%%%%%%%%%%%

%image based vs point cloud based
\par
In the field of medical image analysis, image-based 2D/3D registration is a straightforward method, which directly uses the intraoperative 2D image and preoperative 3D volumetric data as inputs~\cite{hansen2021graphregnet, gao2023fully, lei2024epicardium, Masoumi2021Multimodal, Alessandro2024learning, Ghafurian20173d}. However, 2D slices often lack the capability to represent complex anatomic geometries and suffer from noise~\cite{ferrante2017slice}. Therefore, point cloud registration can be considered a promising alternative, as it reduces data complexity, enhances 3D precision, and ensures compatibility across different imaging modalities~\cite{jauer2018efficient}. In addition, registration in CAS applications often requires surface matching rather than registration between entire images, such as in orthopedic surgery. To this end, the point cloud registration is more direct, while the image-based method will require additional effort to force the algorithm to focus only on the surface~\cite{min2019robust}. Point cloud registration is especially useful when only partial data is available, where the obtained images cannot capture the entire anatomy. Fig.~\ref{Fig_teaser} illustrates an example from a thoracic surgical scenario, where an intra-operative partial point cloud of thoracic cartilage is registered to a complete pre-operative template of the same structure.

\par
Due to the rapid advancements in 3D point representation learning and differentiable optimization, point cloud registration has attracted increasing attention in computer vision. Traditional methods, such as the Iterative Closest Point (ICP) algorithm~\cite{besl1992method} and its variants~\cite{serafin2015nicp, hermans2011robust, jiang2023skeleton}, rely on calculating point correspondences by identifying the nearest point pairs within a defined feature space. The transformation matrix is then optimized based on these correspondences between the source and target point clouds. Such iterative methods are often sensitive to the initial alignment because of the local optima. 

Recent advances in deep learning for point set representation have led to significant progress in extracting features directly from 3D data using Convolutional Neural Networks (CNNs)~\cite{bai2020d3feat, choy2019fully}. A pioneering work, PointNet~\cite{qi2017pointnet}, demonstrated the potential of learning spatial and geometric information directly from raw 3D point clouds. More recently, KPConv~\cite{thomas2019kpconv} has been proposed as an effective and robust method for extracting geometric features, showing strong performance regardless of point cloud density. The extracted point feature can be used to estimate the point correspondence and rigid transformation. However, these methods primarily focus on extracting local features while neglecting the global context of the point cloud, which can cause the registration process to fall into local optima as well—especially when the overlap ratio is low. To address the lack of global context,~\cite{qin2022geometric} proposed a Geometric Transformer where a geometric representation of point clouds is used in the attention computation to aggregate global feature, followed by using this feature in superpoint matching. The transformation-invariant performance is achieved by encoding pair-wise distances and triplet-wise angles as position embedding. However, due to the nature of rotation, it is noncontinuous in the expression in the $3\times3$ rotation matrix. This will result in a decrease in the performance of the rotations that were rarely seen. To address this,~\cite{yu2023rotation} introduced point pair feature-based coordinates to represent pose-invariant geometry in the feature extraction of transformer. However, it uses a pure transformer in an encoder-decoder architecture can lead to higher computational costs and may struggle to capture local features effectively, especially when training data is limited. 

\par
In this study, we propose a learning-based method for point cloud registration under partial overlap, employing a coarse-to-fine strategy that leverages both local and global features for accurate point matching. To effectively capture contextual information at different scales, the encoder combines point convolutions for initial local feature extraction with transformers for further learned feature enhancement on the downsampled point clouds. Additionally, we introduce a novel multi-scale geometry-aware positional embedding within the transformer architecture, designed to enhance geometric perception across varying scales. This embedding is transformation-invariant and robustly preserves the geometric characteristics of individual points.
% In this study, we propose an end-to-end learning-based network for point cloud registration with a partial overlap rate in a coarse-to-fine manner by leveraging local and global features for point matching. To better characterize the local and global contextual information of 3D data respectively, point convolutions and transformers are used in the encoder to capture local contextual features at individual levels and the transformer-based aggregation layer captures the global feature for down-sampled 3D data. Furthermore, we introduce a novel multi-scale geometry-aware feature as positional embedding used in the transformer architecture, aiming to enhance its geometric perception capability across different geometric scales. This positional embedding is transformation-invariant and can robustly preserve the geometric characteristics of individual points in the clouds. 
The key contributions are summarized as follows:
\begin{itemize}
  \item The multi-scale point-wise geometry-aware feature (PGF) representation is proposed to better capture the geometry characteristics for individual points, which is particularly suitable for the partial-to-full registration problem.  
  
  \item The encoder consists of multiple layers, each of which has a KPconv layer for point cloud feature extraction, followed by a transformer layer using PGF as positional embedding for its feature aggregation to enhance its geometry characteristic extraction.

  \item A point cloud registration method featured by its coarse-to-fine structure, in which both local and global information is fused between the partial and full point clouds using a cross-attention mechanism, thereby resulting in robust and accurate performance in partial-to-full registration.

  % \item By introducing the PDNom layer~\cite{wu2024towards}, the proposed method enables a unified model for varying anatomy with significant geometry differences.  
  % % A adaptive representation learning module is used to jointly learn across multiple anatomic structures to enhance the generalization for data from different source. 
  % % \item 
\end{itemize}
To validate the effectiveness of the proposed method, the proposed method is validated on four different bones with significantly varying geometries, namely, tibia, femur, pelvis, and thoracic cartilage. The results demonstrate that the proposed method can outperform existing methods even when the overlap ratio is low or the partial point cloud is not entirely contained in the full point cloud. To encourage reproducibility and facilitate future studies, the code has been released on this webpage
% \footnote{\url{https://github.com/howling1/PGA-for-Partial2Full-Reg}}
.

% \par
% The rest of this paper is structured as follows: Section II discusses related work in more detail. Section III introduces the proposed methods. The experimental results are summarized in Section IV. Finally, the conclusion and discussion are summarized in Section V.

%%%%%%%%%%%%%%%%%%%%%%%%%%%%%%%%%%%%%%%%%%%%%%%%%%%%%%%%%%%%%%%
\section{Related Work}
\subsection{Traditional Point Cloud Registration}
%%%%%%%%%%%%%%ICP
\par
Point cloud registration is a foundational technique in scene understanding within the field of computer vision. One of the most classical approaches is the ICP~\cite{besl1992method}, which iteratively refines the transformation between two point clouds to minimize the distance between corresponding points. To enhance time efficiency and robustness, various ICP variants have been developed. For example, generalized ICP~\cite{segal2009generalized} incorporates a probabilistic framework to better handle noise, while Sparse ICP~\cite{bouaziz2013sparse} emphasizes local patches with higher similarity scores by assigning them greater weight during the registration process. However, such methods are often sensitive to initial alignment due to local optima~\cite{jiang2022precise}. Additionally, high levels of noise will exacerbate these issues by introducing spurious points that mislead the registration process~\cite{maintz1998survey, audette2000algorithmic}.

%%% feature based
\par
Besides ICP and its variants,~\cite{rusu2009fast} proposed to use point feature histograms as multi-dimensional features to describe local geometry for 3D data matching. Similarly,~\cite{frome2004recognizing} introduced 3D shape context and harmonic shape context to capture local shape characteristics. These feature-based methods enable registration by leveraging distinctive geometric descriptions, proving effective when the data exhibits sufficient and unique shapes. However, their performance can significantly degrade when features lack distinctiveness due to noise or low overlap. Moreover, the computation of such features is often time-intensive~\cite{holden2008review, makela2002review}. To improve robustness, the Coherent Point Drift (CPD) algorithm~\cite{myronenko2010point} addresses the registration problem through probabilistic density estimation. However, CPD typically assumes that both the source and target point clouds are complete, a condition that is often difficult to meet in computer-assisted surgery (CAS). In CAS, intra-operative observations are usually only partial compared to the more comprehensive preoperative images, posing a significant challenge for effective registration.

%%%%%%%%%%%%%%%%%%%%%%%%%%%%%%%%%%%%%%%%%%
\subsection{Learning-Based Point Cloud Registration}
\par
Compared to traditional methods, learning-based approaches relied on learning-based point feature extractor to offer more effective and robust feature extraction directly from 3D data for registration. To address the challenges posed by noisy, low-resolution, and incomplete 3D data,~\cite{zeng20173dmatch} proposed a data-driven approach that learns a local volumetric patch descriptor for establishing correspondences between partial 3D datasets. This method leverages CNN for feature extraction, demonstrating significant performance improvements over traditional baselines. To further enhance feature extraction capabilities, various CNN-based encoders have been introduced, such as O-CNN~\cite{wang2017cnn} and FCGF~\cite{choy2019fully}. However, these methods require voxelization of the point clouds. To deal with this issue, recent approaches like PointNet~\cite{qi2017pointnet}, DGCNN~\cite{wang2019dynamic}, and KPConv~\cite{thomas2019kpconv} have been developed to operate directly on point clouds. These advanced feature extractor have been widely used to extract geometric feature for point cloud registration networks. Then according to whether point pair correspondences are explicitly computed, registration networks can be broadly categorized into correspondence-based and correspondence-free. 

\par
Correspondence-free approaches typically involve training an end-to-end network to directly regress the transformation between point clouds. These methods learn a global feature representation of the point clouds and use this global feature to predict the transformation. Approaches such as PointNetLK~\cite{aoki2019pointnetlk}, Featuremetric Registration~\cite{huang2020feature}, and DeepGMR~\cite{yuan2020deepgmr} are computationally efficient since they bypass the need for explicit point-to-point matching. However, while these methods perform well in object-level registration tasks, they often struggle in more complex registration scenarios, where their performance can be suboptimal~\cite{huang2021predator}.

%%%%%%%%%%%%%%%%%%%%%%%%%
\begin{figure*}[ht!]
\centering
\includegraphics[width=0.85\textwidth]{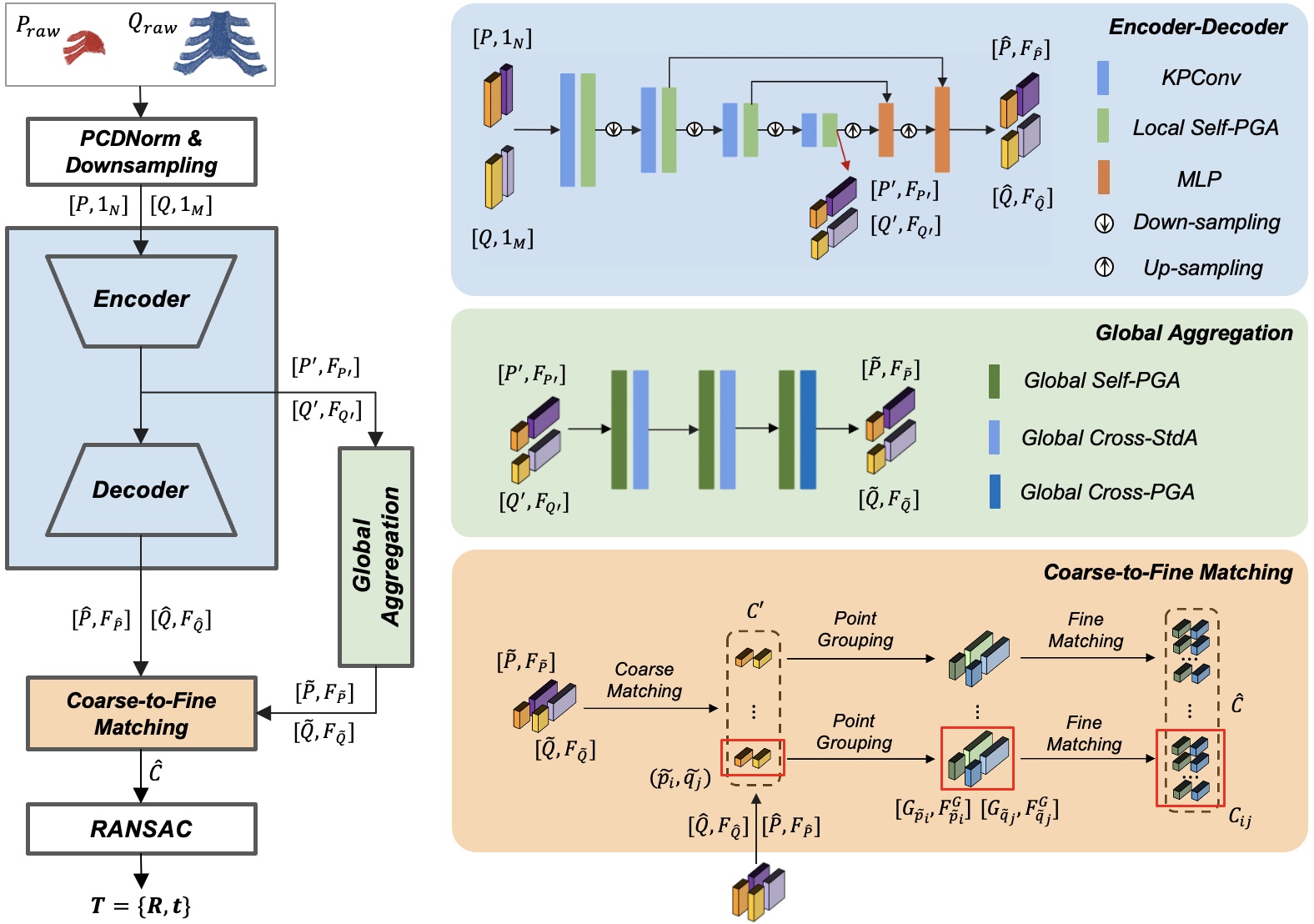}
\caption{The workflow of the proposed \emph{GAPR-Net} is illustrated as follows. The encoder in the backbone extracts local features from $\mathbf{P}$ and $\mathbf{Q}$ and downsamples them into keypoints $\mathbf{P'}$ and $\mathbf{Q'}$. The decoder then upsamples $\mathbf{P'}$ and $\mathbf{Q'}$ to generate fine-grained point clouds $\mathbf{\hat{P}}$ and $\mathbf{\hat{Q}}$. A global aggregation module learns and integrates global contextual information from the downsampled $\mathbf{P'}$ and $\mathbf{Q'}$, producing updated point features for $\mathbf{\tilde{P}}$ and $\mathbf{\tilde{Q}}$ ($\mathbf{\tilde{P}}=\mathbf{P'}$, $\mathbf{\tilde{Q}}=\mathbf{Q'}$). These, along with the fine-grained point sets $\mathbf{\hat{P}}$ and $\mathbf{\hat{Q}}$, are used in a coarse-to-fine matching scheme to estimate the point correspondences $\mathbf{\hat{C}}$ between $\mathbf{\hat{P}}$ and $\mathbf{\hat{Q}}$. Finally, RANSAC is applied to the correspondences $\mathbf{\hat{C}}$ to estimate the rigid transformation $\mathbf{T}$. Throughout the above process, the network only updates the point features of the point clouds. The spatial coordinates of the point clouds themselves are only affected by the downsampling and upsampling operations.}
\label{Fig_workflow}
\end{figure*}
%%%%%%%%%%%%%%%%%%%%%%%%%

\par
Correspondence-based methods focus on identifying pairs of corresponding points between point clouds. The rigid transformation is obtained via non-learning process such as RANSAC~\cite{shen2020ransac}, SVD
decomposition~\cite{eckart1936approximation}. The correspondence-based approaches can be further categorized into patch-wise and point-wise approaches. Patch-wise methods, such as Lepard~\cite{li2022lepard}, Deep Closest Point (DCP)~\cite{li2019deep}, and RORNet~\cite{shen2020ransac}, represent a significant subset of this category. These methods typically employ a shared encoder to extract features from the input point clouds while simultaneously performing geometric downsampling. Common techniques for this downsampling include grid downsampling~\cite{rusu20113d} and Farthest Point Sampling (FPS)~\cite{qi2017pointnet++}. The encoder generates key points that encapsulate the local geometric features of patches. By matching these key points, the transformation between point clouds can be estimated. This patch-wise approach reduces the computational burden associated with point-wise matching, leading to a more efficient registration process. Unlike patch-wise methods, point-wise approaches focus on learning dense features for all or most points in the point cloud, considering potential matches for each point. While this approach results in higher time and space complexity, it avoids the accuracy loss associated with keypoint extraction in patch-wise methods. Predator~\cite{huang2021predator} is a pioneering work in this domain, employing a KPConv-based backbone to extract point features for precise matching. CoFiNet~\cite{yu2021cofinet} introduced a coarse-to-fine matching scheme that integrates both point-wise and patch-wise techniques, balancing efficiency with high matching accuracy, particularly in scenarios with partial overlap. Building on this approach, GeoTransformer~\cite{qin2022geometric} and RoITr~\cite{yu2023rotation} further advanced the field, achieving SOTA performance. Our method also adopts the coarse-to-fine matching strategy to enhance point-wise matching, ensuring accuracy while improving efficiency.

%%%%%%%%%%%%%%%%%%%%%%%%%%%%%%%%%%%%%%%%%%
\section{Partial and Full Point Clouds Dataset} \label{data-preprocessing}
\par
To develop the partial-to-full registration for modern navigated surgery, particularly in orthopedic applications, we extract the full level point cloud from CTs of multiple types of bone, including pelvis, femur, tibia, and thoracic cartilage. The thoracic cartilage point clouds were manually extracted from the RibFrac dataset~\cite{ribfracchallenge2024,ribfracclinical2020}, whereas the other bone point clouds were sourced from the dataset used in~\cite{chen2023generalized}. The final dataset consists of 202 pelvis point clouds, 358 femur point clouds, 397 tibia point clouds, and 100 thoracic cartilage point clouds.

% Bone shape registration is a common scenario in CAS, with applications such as robotic surgery. We evaluate \emph{GAPR-Net} on an open-source, multi-class bone point cloud dataset derived from CT scans~\cite{chen2023generalized}, which includes point cloud data of the pelvis, femur, and tibia. Additionally, we follow the pipeline of \cite{jiang2023skeleton,jiang2023thoracic} to manually create a thoracic cartilage point cloud dataset consisting of 100 samples from RibFrac\cite{ribfracchallenge2024,ribfracclinical2020}, which is an open-sourced dataset of upper-body CT scan, as a supplement. 

\par
To generate the paired partial point cloud, a point is randomly selected from the full point cloud (for tibia and femur, the point is chosen from either end to ensure sufficient geometric features), and its $K$-nearest neighbors (KNN) are retained to preserve a predefined proportion of the original points. This proportion is randomly sampled between 40\% and 70\%. Given that partial observations may originate from accessible yet low-quality images, noise and artifacts can affect both the point density and spatial accuracy in the resulting point cloud. To simulate these effects, the partial point cloud is further perturbed by introducing randomness in point density and local point positions.

\par
We first apply the point set normalization method from PointNet~\cite{qi2017pointnet} to center the partial point cloud at the origin $(0, 0, 0)$ and scale the longest axis among the XYZ dimensions to the range $[-1, 1]$. To simulate varying point densities, a point is randomly selected from the normalized partial point cloud, and local up-sampling or down-sampling is performed within its neighborhood, where the neighborhood radius is randomly chosen between 0.1\,mm and 0.15\,mm. Down-sampling is performed by randomly discarding points, whereas up-sampling is achieved by generating interpolated points. Specifically, for up-sampling, two points $p_1$ and $p_2$ are randomly selected within the local neighborhood, and a new point is computed as $p_{\text{new}} = \alpha \cdot p_1 + (1 - \alpha) \cdot p_2$, where $\alpha$ is a random scalar uniformly sampled from $[0,1]$. This process is repeated multiple times to induce local variations in point density. To further simulate the point location variation between partial and full point clouds, Gaussian noise is independently added to each point coordinate. The noise is sampled from a zero-mean Gaussian distribution $\mathcal{N}(0, \sigma^2)$, where $\sigma$ is set to 0.01. The partial point cloud is then denormalized to its original scale and location. Finally, a random rigid transformation $\mathbf{T} = \{\mathbf{R}, \mathbf{t}\}$ is applied to the partial point cloud. The rotation matrix $\mathbf{R}$ represents rotations around the X, Y, and Z axes, each independently sampled from a uniform distribution over $[0^\circ, 180^\circ]$. The translation vector $\mathbf{t}$ consists of values randomly sampled within the range $[-150\,\mathrm{mm},\ +150\,\mathrm{mm}]$ along corresponding axis. By repeating the aforementioned process, we generated 2314 unique pairs of partial and full point clouds. The dataset comprises 400, 404, 716, and 794 point cloud pairs for thoracic cartilage, pelvis, femur, and tibia, respectively. In this study, 60\% of the data for each bone type is allocated for training, 20\% for validation, and the remaining 20\% for testing.

\section{Methods}
Considering the strict requirements of CAS applications, we also follow the correspondence-based method to ensure the robustness of the registration. Given a partial point cloud $\mathbf{P_{raw}} = \{ p_i \in \mathbb{R}^3 \mid i = 1, \ldots, N_{raw} \}$ representing the intro-operative partial observation, and a full point cloud $\mathbf{Q_{raw}} = \{ q_i \in \mathbb{R}^3 \mid i = 1, \ldots, M_{raw} \}$ obtained in pre-operative phase for registration, our registration framework first predicts point correspondences $\mathbf{\hat{C}}$ using a neural network architecture termed the Geometry-Aware Partial-to-Full Registration Network (\emph{GAPR-Net}). Subsequently, the corresponding rigid transformation $\mathbf{T} = \{\mathbf{R}, \mathbf{t}\}$ is computed by RANSAC~\cite{fischler1981random}, where $\mathbf{R} \in SO(3)$ is the 3D rotation and $\mathbf{t} \in \mathbb{R}^3$ is the 3D translation.

\par
The proposed \emph{GAPR-Net} consists of a shared encoder-decoder backbone, a transformer-based global aggregation, and a coarse-to-fine matching module. The overall pipeline is illustrated in Fig.~\ref{Fig_workflow}. We begin by applying point cloud normalization~\cite{qi2017pointnet} (PCDNorm) and voxel downsampling to $\mathbf{P_{raw}}$ and $\mathbf{Q_{raw}}$ to standardize scale and point density during training while reducing computational overhead. We then associate the normalized and downsampled clouds $\mathbf{P} \in \mathbb{R}^{N \times 3}$ and $\mathbf{Q} \in \mathbb{R}^{M \times 3}$ with initial point features, where all feature values are initialized to $\mathbf{1}$. Specifically, the inputs $[\mathbf{P}, \mathbf{1_{|P|}}]$ and $[\mathbf{Q}, \mathbf{1_{|Q|}}]$, where $\mathbf{1}_{*}$ denotes an all-one vector of length $*$, are fed into a shared encoder-decoder network to extract local point-wise features. The encoder-decoder structure is built upon the KPConv-FPN backbone~\cite{thomas2019kpconv, qin2022geometric} to capture the multi-scale features for the point clouds. The encoder has four layers, including a bottleneck one. To extract abstract representations from the input point clouds, each layer in the encoder first applies KPConv to compute features for each point, followed by a local self-attention mechanism to aggregate contextual information from neighboring points. 

% To improve computational efficiency and enhance keypoint representations, a downsampling operation is subsequently applied—except at the bottleneck layer—to progressively reduce the number of points.

Then, the downsampled point clouds and their corresponding features, $[\mathbf{P'}, \mathbf{F_{P'}}]$ and $[\mathbf{Q'}, \mathbf{F_{Q'}}]$, at the bottleneck are fed into a transformer-based global aggregation module to capture global contextual information. To facilitate globally optimal alignment between the partial and full point clouds, self-attention and cross-attention mechanisms with a larger perception field are applied in an alternating fashion, allowing the network to explicitly relate and compare features from the two point clouds. In this study, the attention block is repeated three times to perform coarse matching. Prior to the final cross-attention, an additional refinement stage is introduced, where the proposed geometry-aware feature encoding is incorporated as a positional embedding in the subsequent cross-attention to improve correspondence estimation accuracy. Notably, the first two cross-attention modules adopt the standard vanilla transformer design, as the inclusion of additional positional embeddings was observed to hinder training convergence, consistent with the findings in~\cite{qin2022geometric}. To enhance registration accuracy at a dense level, the upsampled point clouds and their corresponding features from the decoder, i.e., $[\mathbf{\hat{P}}, \mathbf{F_{\hat{P}}}]$ and $[\mathbf{\hat{Q}}, \mathbf{F_{\hat{Q}}}]$, together with the coarse matching results, are utilized to perform fine-grained matching within each corresponding patch. In this way, the proposed method leverages both global cues to improve registration robustness and local point-wise features to ensure fine-grained matching accuracy at the dense point cloud level. The details of each module are provided in the following subsections.

%%%%%%%%%%%%%%%%%%%%%%%%%%%%%%%%%%%%%%%%%%%%%%%%%%%%%
\subsection{Point-Wise Geometry-Aware Feature Representation}
\label{attention feature}
%%%%%%%%%%%%%%%%%%%%%%%%%
\begin{figure}[ht!]
\centering
\includegraphics[width=0.40\textwidth]{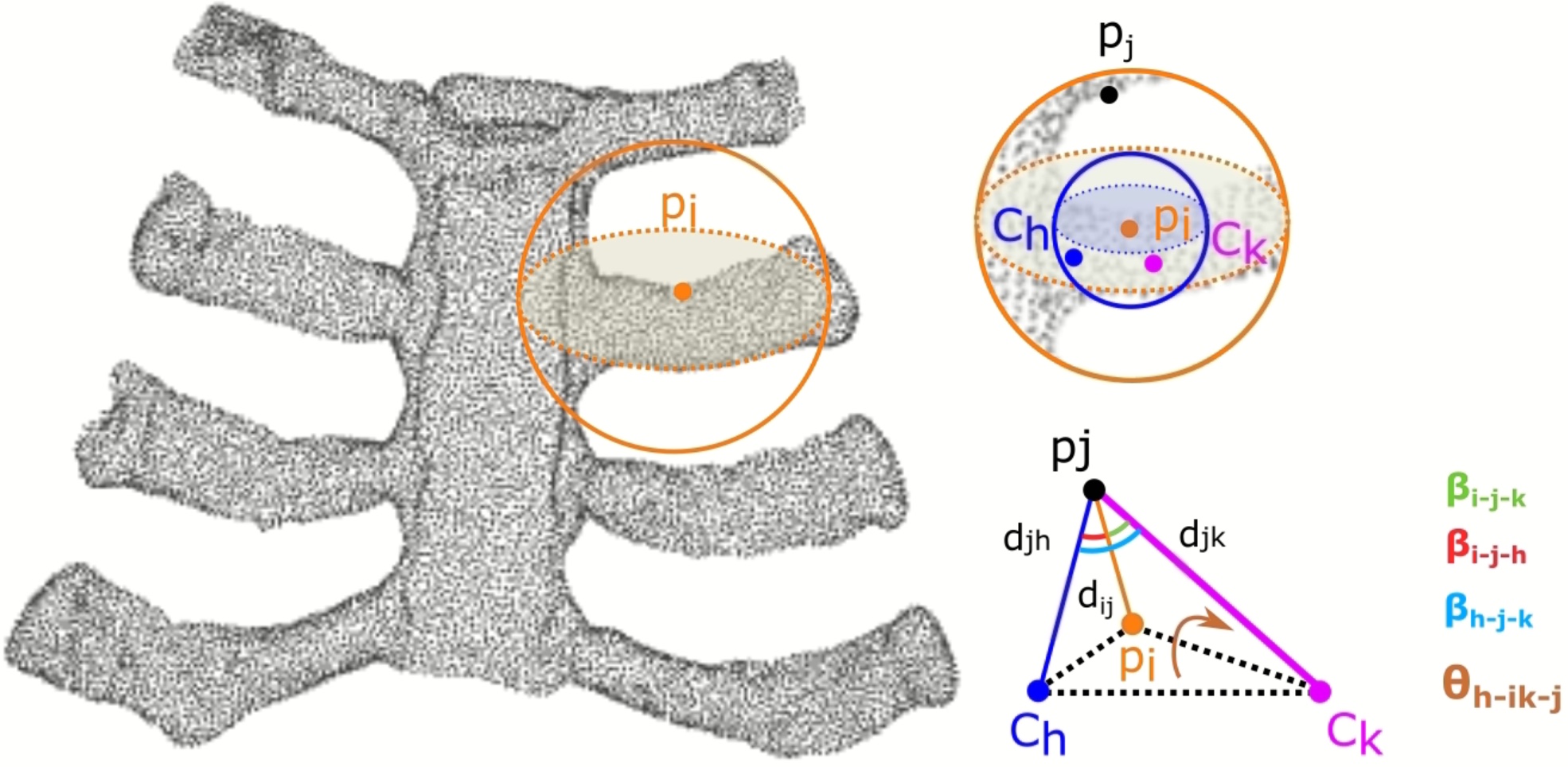}
\caption{The extraction of the \emph{PGF} for $p_j$ with regard to $p_i$. The orange ball and blue ball represent the first larger neighborhood of $p_i$ and the second smaller neighborhood $p_i$ respectively. The blue, magenta and orange edge line represent $d_{jh}$, $d_{ik}$, $d_{ij}$. The green, red, blue and brown angles represent the $\beta_{i-j-k}$, $\beta_{i-j-h}$, $\beta_{h-j-k}$, $\theta_{h-ik-j}$ respectively.}
\label{Fig_position_rep}
\end{figure}
%%%%%%%%%%%%%%%%%%%%%%%%%

\par
To perform point cloud registration or matching, it is crucial to effectively characterize the geometric features of individual points within a 3D point cloud. To ensure consistent representation, the features should remain invariant to both translation and rotation. Inspired by the existing works for geometric features encoding of the 3D point cloud~\cite{weinmann2017geometric, weinmann2015semantic, li2021rotation}, we present a multi-scale Point-Wise Geometry-Aware Feature (\emph{PGF}), consisting of seven variables $l_{ij} = (d_{ij}, d_{jh}, d_{jk}, \beta_{i-j-k}, \beta_{i-j-h}, \beta_{h-j-k}, \theta_{h-ik-j})$. The feature describes the relative positional features of individual $p_j$ with regard to individual $p_i$ in the point cloud. This geometric feature is specifically designed for partial-to-full registration, where the centroid of the full and partial point clouds literally will not overlap in most cases after registration. The \emph{PGF} serves as the foundation for attention computation in our method.

\par
An illustration of \emph{PGF} for the point cloud $\mathbf{P}$ is provided in Fig.~\ref{Fig_position_rep}. For each point $p_{i} \in \mathbf{P}$, a query ball with radius $r$ (illustrated as the orange sphere in Fig.~\ref{Fig_position_rep}) is centered at $p_i$ to define its local neighborhood. For any neighboring point $p_j$ within this region, we seek features that capture the relative positional information between $p_j$ and $p_i$. As the first step, the centroid $c_h$ of the query ball is computed. Since three points can only describe the feature in a plane, the spatial information for a 3D point cloud is still lacking. We define a second query ball centered at $p_i$ with a radius of $k_r \cdot r$, where $k_r$ is a hyperparameter less than 1.0. The centroid of points in the second searching ball is computed as the fourth feature point $c_k$. This combination of small and large neighborhoods facilitates the extraction of geometric features at multiple scales. To capture the unique representation of each point in 3D space, the final geometric feature $l_{ij}$ is defined as seven distinctive variables derived from the four feature points as follows: \\ 
(i) $d_{ij}$: the distance from $p_{i}$ to $p_j$; \\
(ii) $d_{jh}$: the distance from $p_{j}$ to $c_h$;\\
(iii) $d_{jk}$: the distance from $p_{j}$ to $c_k$; \\ 
(iv) $\beta_{i-j-k}$: the angle between $p_i-p_j$ and $c_k-p_j$; \\
(v) $\beta_{i-j-h}$: the angle between $p_i-p_j$ and $c_h-p_j$; \\
(vi) $\beta_{h-j-k}$: the angle between $c_h-p_j$ and $c_k-p_j$;  \\
(vii) $\theta_{h-ik-j}$: the angle between the plane defined by the triangle $c_h$-$p_i$-$c_k$ and the plane defined by the triangle $p_j$-$c_k$-$p_i$, measured around the line $p_i$-$c_k$.

%%%%%%%%%%%%%%%%%%%%%%%%%%%%%%%%%%%%%%%%%%%%%%%%%%%%
\subsection{Point-Wise Geometry-Aware Self-Attention Computation}
\label{self-attn section}
% \todo{1.conditional pe?}

%%%%%%%%%%%%%%%%%%%%%%%%%
\begin{figure}[ht!]
\centering
\includegraphics[width=0.30\textwidth]{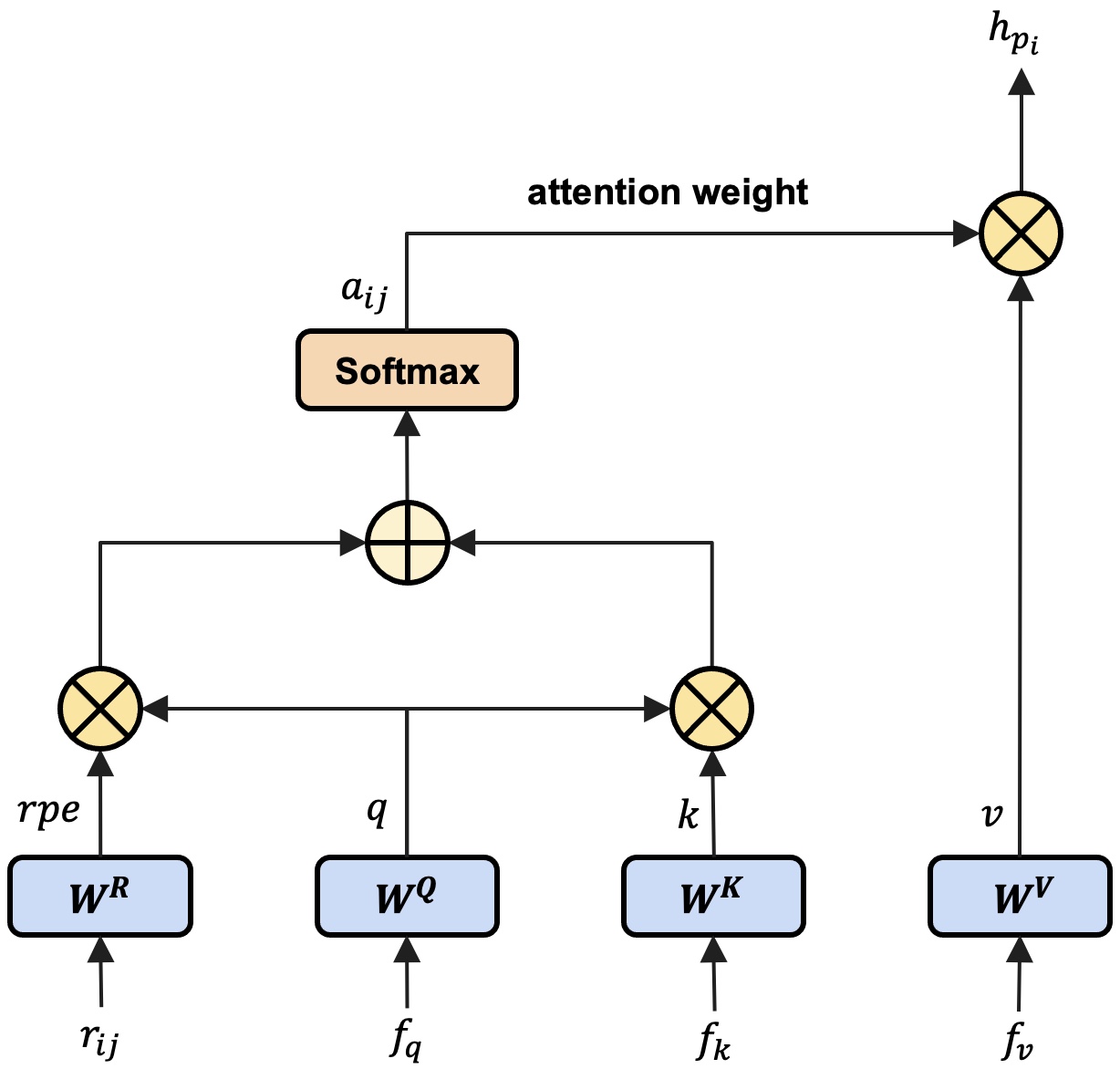}
\caption{Attention computation for Point-Wise Geometry-Aware Attention.}
\label{Fig_attn_comp}
\end{figure}
%%%%%%%%%%%%%%%%%%%%%%%%%

\par
In the encoder, the KPConv~\cite{thomas2019kpconv} is the state-of-the-art method to turn the processing of point clouds into a similar way to CNN for images, enabling the design of deep structure. For the partial point cloud $\mathbf{P}$, after the KPConv process, point features $\mathbf{F_P}$ represents points' context information in 3D for $\mathbf{P}$. To further aggregate this feature to enhance the structural understanding, a self-attention transformer layer is applied after each KPConv layer. To fuse the understanding of geometry into the network, a Point-Wise Geometry-Aware Attention (\emph{PGA}) mechanism is bulit based on the \emph{PGF} for self-attention compuatation. The detailed computation of \emph{PGA} is demonstrated in Fig.~\ref{Fig_attn_comp}. In the \emph{PGA}-based self-attention (\emph{self-PGA}), for each point $p_i \in \mathbf{P}$, a fixed radius $r$—identical to that used in the preceding KPConv neighborhood search—is employed to identify its neighboring points. Then, the geometry-aware feature $l_{ij}$ is computed between $p_i$ and each neighboring point $p_j$. Since the context is restricted to the local neighborhood of each point, we refer to the \emph{PGA} used in the encoder as local \emph{PGA}. In contrast, the context in the \emph{PGA} of the global aggregation module spans the entire point cloud. To distinguish it from the local version, we refer to the \emph{PGA} in the global aggregation module as global \emph{PGA}.

\par
In the \emph{self-PGA}, the relative positional embedding (RPE) $r_{ij} \in \mathbb{R}^d$ and the query feature $f_q \in \mathbb{R}^d$ are obtained via separate MLP projections applied to the geometry-aware feature $l_{ij}$ and $p_i$'s point feature $f_{p_i}$, respectively. The key $f_k \in \mathbb{R}^d$ and value $f_v \in \mathbb{R}^d$ are computed by applying individual MLP projections to the feature of each neighboring point $p_j$, i.e., $f_{p_j}$. The updated output feature $h_{p_i} \in \mathbb{R}^d$ for point $p_i$ is then computed as:

\begin{equation}
h_{p_i} = \sum_{j=1}^{|\mathcal{N}(p_i)|} a_{ij} (f_v W^V),
\end{equation}
where $\mathcal{N}(p_i)$ represents the number of neighboring points of $p_i$, $W^V \in \mathbb{R}^{d \times d}$ is the learnable projection matrix applied to the value, and $a_{ij}$ is the attention weight between $p_i$ and $p_j$, computed as:
\begin{equation}
a_{ij} = \text{softmax}\left(\frac{(f_q W^Q) \left( f_k W^K + r_{ij} W^R \right)^T }{\sqrt{d}}\right).
\label{eq:attn}
\end{equation}
Here, the matrices $W^Q$, $W^K$, $W^V$, and $W^R \in \mathbb{R}^{d \times d}$ are the learnable projection matrices for the query, key, value, and RPE, respectively.

%%%%%%%%%%%%%%%%%%%%%%%%%%%%%%%%%%%%%%%%%%%%%%%%%%%%%%%%%%%%%%%%%%%%
% \subsection{Aggregator based on Multi-Scale Geometric Transformer}
\subsection{Global Feature Aggregation}
\par
While the backbone effectively captures local features from individual points, incorporating global context is also essential for accurate registration. Additionally, effective feature exchange between the partial and full point clouds is critical for establishing reliable point-to-point correspondences. To address this issue, we introduce the global \emph{PGA} for the self-attention in the global aggregation module. Specifically, in the computation of the \emph{PGF}, the larger neighborhood is defined as the entire point cloud, while the smaller neighborhood uses the same radius as the KPConv neighborhood ball from the encoder bottleneck layer. Furthermore, to enhance alignment between partial and full point clouds, a global cross-attention feature aggregation module is applied after each global \emph{self-PGA}. Here, \emph{global} denotes that all points from the counterpart cloud participate in the attention computation. The global feature aggregation module adopts a stacked architecture of global self-cross-PGA layers, where the number of stacked layers is set to 3 in our study, as illustrated in Fig.~\ref{Fig_workflow}.

It is important to note that the initial two cross-attention modules follows the standard vanilla transformer structure and does not include the relative positional embedding defined in this study. This design is based on the observation that introducing positional embeddings in an early phase can hinder training convergence, which is consistent with findings reported in~\cite{qin2022geometric}. This is because such embedding cues are only effective for aligning features once the point clouds are coarsely registered. In other words, if a rough repositioning is first applied to one of the input point clouds to achieve approximate alignment, the positional embeddings can then provide meaningful guidance~\cite{li2022lepard}. Motivated by this insight, we estimate a rough transformation using the output of the second standard cross-attention module, which is applied to align one point cloud to the other. Based on this coarse alignment, we then introduce the \emph{cross-PGA} into the final cross-attention layer within the global aggregation module. This allows the geometry-aware feature $l_{ij}$ to better capture meaningful relative positional information, thereby facilitating more accurate and fine-grained registration.

\par
For the \emph{cross-PGA}, the computation follows the same structure as illustrated in Fig.~\ref{Fig_attn_comp}. Specifically, the \emph{PGF} $l_{ij}$ is first computed across the point clouds, where the entire counterpart point cloud serves as the larger neighborhood, and the smaller neighborhood is defined by the KPConv neighborhood ball at the encoder bottleneck layer, consistent with the configuration used in the global \emph{self-PGA}. Then the relative positional embedding $r_{ij} \in \mathbb{R}^{d'}$ and the query feature $f_q \in \mathbb{R}^{d'}$ are obtained via separate MLP projections applied to the geometry-aware feature $l_{ij}$ and the input point feature of $p'_i$ (i.e., $f_{p'_i}$). The key $f_k \in \mathbb{R}^{d'}$ and the value $f_v \in \mathbb{R}^{d'}$ are computed by applying separate MLP projections to the point feature $f_{q'_j}$ from the counterpart point cloud $\mathbf{Q'}$. Then, the output point feature $h_{p'_i}^{*} \in \mathbb{R}^{d'}$ for $p'_i$ is computed as:
\begin{equation}
h_{p'_i}^{*} = \sum_{j=1}^{|\mathbf{Q'}|} a_{ij}^{*} (f_v W^V),
\label{eq:cross-feats}
\end{equation}
indicating that the features of all points in the counterpart point cloud $\mathbf{Q'}$ are taken into account. Here, $^{*} \in \{\text{std}, \text{\emph{PGA}}\}$ denotes whether standard attention or \emph{PGA} is employed, $a_{ij}^{*}$ is the attention weight. 

For the standard attention used in the first two cross-attention modules, the attention weights are computed as:
\begin{equation}
a_{ij}^{\text{std}} = \text{softmax}\left(\frac{(f_q W^Q) (f_k W^K)^T}{\sqrt{d'}}\right).
\label{eq_cross_att_std}
\end{equation}
For the final cross-attention block that incorporates \emph{cross-PGA}, a repositioning step is first conducted. Specifically, the output features from the preceding self-attention module are used to compute the soft point correspondences $\mathbf{C_r}$ (see Sec.~\ref{sec:coarse matching}). A weighted SVD decomposition~\cite{arun1987least} based on $\mathbf{C_r}$ is then performed to estimate the transformation $T_r$, which is applied to update the coordinates of $\mathbf{Q'}$. Using the repositioned $\mathbf{Q'}$, the attention weights for \emph{cross-PGA} are computed as:
\begin{equation}
a_{ij}^{\emph{PGA}} = \text{softmax}\left(\frac{(f_q W^Q) \left(f_k W^K + r_{ij} W^R\right)^T}{\sqrt{d'}}\right).
\label{eq_cross_att_PGCA}
\end{equation}

%%%%%%%%%%%%%%%%%%%%%%%%%%%%%%%%%%%%%%%%%%%
\subsection{Coarse-to-Fine Matching}
\par
Due to the partial overlap between the partial point cloud \(\mathbf{P}\) and the full point cloud \(\mathbf{Q}\), computing matching probabilities for all possible point pairs is both inefficient and unnecessary. To address this, we adopt the coarse-to-fine matching strategy proposed in CoFiNet~\cite{yu2021cofinet}, which significantly improves efficiency in partial point cloud registration.

\par
To enable coarse-to-fine matching, we first employ a point-to-node grouping strategy to construct local patches centered around each downsampled point $\tilde{p}_i \in \mathbf{\tilde{P}}$ obtained from the global aggregation module. Each fine-grained point $\hat{p}_i \in \mathbf{\hat{P}}$ from the decoder is assigned to its nearest downsampled point $\tilde{p}_i$ based on the Euclidean distance. The resulting local patch centered at $\tilde{p}_i$ is denoted as $G_{\tilde{p}_i}$. To ensure the validity of correspondence estimation, any downsampled point that does not receive an assignment is discarded. The same procedure is symmetrically applied to the full point cloud \( \mathbf{\tilde{Q}} \), yielding \( G_{\tilde{q}_j} \) for each downsampled point \( \tilde{q}_j \).

% \begin{equation}
% G_{p_i} = \{\hat{p_i} \in \hat{P} \mid i = \arg\min_j (|| \hat{p_i} - \tilde{p}_j ||_2^2), \tilde{p}_j \in \tilde{P} \}
% \end{equation}

\subsubsection{Coarse Matching}
\label{sec:coarse matching}
During the coarse matching stage, correspondences \( \mathbf{C'} \) are established between the downsampled point clouds \(\mathbf{\tilde{P}}\) and \(\mathbf{\tilde{Q}}\) produced by the global aggregation module. Specifically, we first construct a Gaussian correlation matrix \( S \in \mathbb{R}^\mathbf{{|\tilde{P}|} \times \mathbf{|\tilde{Q}|}} \), where each element \( s_{ij} \) is defined as:
\begin{equation}
s_{ij} = \exp(-||f_{\tilde{p}_i} - f_{\tilde{q}_j}||_2^2).
\end{equation}
Here, \( f_{\tilde{p}_i} \) and \( f_{\tilde{q}_j} \) denote the point features of $\tilde{p}_i \in \tilde{P}$ and $\tilde{q}_j \in \tilde{Q}$, respectively. To suppress ambiguous correspondences, we apply dual-normalization~\cite{aboyans20182017} to the matrix \( S \), resulting in a normalized correlation matrix \( \bar{S} \) computed as:
\begin{equation}
\bar{s}_{ij} = \frac{s_{ij}}{\sum_{k=1}^{|\mathbf{\tilde{Q}}|} s_{ik}} \cdot \frac{s_{ij}}{\sum_{k=1}^{|\mathbf{\tilde{P}}|} s_{kj}}.
\end{equation}
Finally, we select the top-\(k\) highest-scoring entries from \( \bar{S} \) as the correspondences \( C' \) for the downsampled point clouds \( \mathbf{\tilde{P}} \) and \(\mathbf{ \tilde{Q}} \).

\subsubsection{Fine Matching}
The downsampled correspondences \( C' \) inherently represent patch-level correspondences. In the fine matching stage, we restrict the search for fine-grained correspondences to those within the matched patch pairs in \( C' \). For each correspondence \( C'_{ij} = (\tilde{p}_i, \tilde{q}_j) \in C' \), we apply an optimal transport layer~\cite{sarlin2020superglue} to compute fine-grained correspondences \( C_{ij} \) between the local patches \( [G_{\tilde{p}_i}, F^G_{\tilde{p}_i}] \) and \( [G_{\tilde{q}_j}, F^G_{\tilde{q}_j}] \), where \( F^G_{\tilde{p}_i} \) and \( F^G_{\tilde{q}_j} \) denote the respective point feature sets. Specifically, we begin by computing a cost matrix \( \text{Cost}_{ij} \) for the fine-grained points in \( G_{\tilde{p}_i} \) and \( G_{\tilde{q}_i} \) as follows:
\begin{equation}
\text{Cost}_{ij} = \frac{F^G_{\tilde{p}_i} (F^G_{\tilde{q}_j})^T}{\sqrt{\hat{d}}},
\end{equation}
where \( \hat{d} \) is the corresponding feature dimension. Following the SuperGlue~\cite{sarlin2020superglue}, we augment the cost matrix \( \text{Cost}_{ij} \) by appending a new row and column, both filled with a learnable dustbin parameter \( \alpha \), resulting in the augmented cost matrix \( \bar{C}_{ij} \). The Sinkhorn algorithm~\cite{sinkhorn1967concerning} is then applied to this augmented matrix to obtain a doubly stochastic matrix \( \bar{Z}_{ij} \). After removing the appended row and column, we obtain the soft correspondences \( Z_{ij} \). From \( Z_{ij} \), we extract the point matches corresponding to the top-\(k\) values in both their respective rows and columns. The resulting fine-grained correspondences for each patch pair \( (G_{\tilde{p}_i}, G_{\tilde{q}_i}) \) are denoted as \( C_{ij} \). Finally, we aggregate all the fine-grained correspondences to obtain the full set of fine-grained point correspondences:
\begin{equation}
\mathbf{\hat{C}} = \bigcup_{(\tilde{p}_i, \tilde{q}_j) \in C'} C_{ij}.
\end{equation}
The fine-grained correspondences $\mathbf{\hat{C}}$ are then used to optimize the final registration matrix via the RANSAC algorithm~\cite{shen2020ransac}.

\subsection{Loss Function}
\par
In the training, we use separate loss terms for coarse matching ($\mathcal{L}_{cm}$) and fine matching ($\mathcal{L}_{fm}$) to optimize both processes simultaneously. The total loss is given by:
\begin{equation}
\mathcal{L} = \mathcal{L}_{cm} + \mathcal{L}_{fm}.
\end{equation}

\subsubsection{Coarse Matching Loss}
For the coarse matching loss \( \mathcal{L}_{cm} \), we adopt the overlap-aware circle loss proposed in GeoTransformer~\cite{qin2022geometric}, computed separately for the partial point cloud \( \mathbf{P} \) and the full point cloud \( \mathbf{Q} \). The total coarse matching loss is defined as:
\begin{equation}
\mathcal{L}_{cm} = \frac{\mathcal{L}_{cm}^P + \mathcal{L}_{cm}^Q}{2}.
\end{equation}

We detail the computation of \( \mathcal{L}_{cm}^P \) below; the loss \( \mathcal{L}_{cm}^Q \) is computed analogously. We first define patch-level correspondence criteria: if two patches have at least 10\% overlap, they are considered a positive pair; if they have no overlap, they form a negative pair. Based on this criterion, we construct a patch set \( A \) for point cloud \( \mathbf{P} \), where each patch \( G_{\tilde{p}_i} \in A \) has at least one positive match in \( \mathbf{Q} \). For each \( G_{\tilde{p}_i} \in A \), let \( \varepsilon_{p_i} \) and \( \varepsilon_{n_i} \) denote its sets of positive and negative patches in \( \mathbf{Q} \), respectively. The overlap-aware circle loss for \( \mathbf{P} \) is then given by:
\begin{multline}
\mathcal{L}_{cm}^P = \frac{1}{|A|} \sum_{G_{\tilde{p}_i} \in A} \log \left[ 1 +
\sum_{G_{\tilde{q}_j} \in \varepsilon_{p_i}} e^{\lambda_i^j \beta^{i,j}_p (d_i^j - \Delta_p)}
\right. \\
\left. \times \sum_{G_{\tilde{q}_k} \in \varepsilon_{n_i}} e^{\beta^{i,k}_n (\Delta_n - d_i^k)} \right].
\end{multline}
Here, $d_i^j = || f_{\tilde{p}_i} - f_{\tilde{q}_i} ||_2$ and $d_i^k = || f_{\tilde{p}_i} - f_{\tilde{q}_k} ||_2$ denote the Euclidean distances between the feature embeddings of the corresponding downsampled point pairs. The weighting factor $\lambda_i^j = (o_i^j)^{1/2}$ is introduced to dynamically scale the loss according to the overlap ratio $o_i^j$ between patches $G_{\tilde{p}_i}$ and $G_{\tilde{q}_i}$, thereby encouraging the model to focus more on patch pairs with higher geometric overlap. The positive and negative sample weights are defined as \( \beta^{i,j}_p = \gamma(d_i^j - \Delta_p) \) and \( \beta^{i,k}_n = \gamma(\Delta_n - d_i^k) \), where \( \gamma(\cdot) \) is the element-wise ReLU function. The hyperparameters \( \Delta_p = 0.1 \) and \( \Delta_n = 1.4 \) control the positive and negative margins, respectively, and are set following the recommendations in~\cite{qin2022geometric} to balance the contributions of the positive and negative terms in the loss.

\subsubsection{Fine Matching Loss}
For the fine matching stage, we adopt a negative log-likelihood loss on the doubly stochastic matrix \( \bar{Z}_{ij} \) obtained during fine matching, following the approach in SuperGlue~\cite{sarlin2020superglue}. During training, we randomly sample \( N_c \) ground-truth downsampled correspondences, denoted as \( C^* \). For each matched point pair \( C^*_{ij} \in C^* \), we extract a set of fine-grained ground-truth correspondences \( M_{ij} \) from the associated local patches, where each correspondence is determined using a predefined matching radius (0.05\,mm for normalized point clouds in our study). Fine-grained points in the corresponding local patches that do not form valid matches are grouped into unmatched sets \( I_i \) and \( J_j \) for the two patches, respectively. The fine matching loss for each patch pair \( C^*_{ij} \) is then defined as:
\begin{equation}
\mathcal{L}_f^{ij} = -\sum_{(x, y) \in M_{ij}} \log \bar{z}_{ij}^{xy} - \sum_{x \in I_i} \log \bar{z}_{ij}^{x, m_i+1} - \sum_{y \in J_i} \log \bar{z}_{ij}^{n_i+1, y},
\end{equation}
where \( \bar{z}_{ij}^{xy} \) denotes the soft assignment score in the doubly stochastic matrix $ \bar{Z}_{ij} $ between point $ x \in  G_{\tilde{p}_i} $ and point $ y  \in  G_{\tilde{q}_i} $ obtained during the fine matching stage. The indices \( m_i+1 \) and \( n_i+1 \) represent the dustbin entries corresponding to unmatched points in the respective local patches. Finally, the overall fine matching loss is computed by averaging the losses across all downsampled correspondence pairs:
\begin{equation}
\mathcal{L}_{fm} = \frac{1}{|C^*|} \sum_{C_{ij}^* \in C^*} \mathcal{L}_f^{ij}.
\end{equation}

%%%%%%%%%%%%%%%%%%%%%%%%%%%%%%%%%%%%%%%%%%%%%%%%%%%%%%%%%%%%%%%%%%%%%%%%%%%%%%%%%%%
\section{Results}
\par
This section summarizes the performance of the proposed \emph{GAPR-Net} in comparison with existing methods for partial-to-full point cloud registration (see Sec.~\ref{sec:results_sota}). To further demonstrate the practical effectiveness of our approach, we additionally evaluate its performance on point cloud pairs with varying overlap ratios (see Sec.~\ref{sec:results_overlap}). We also consider more challenging scenarios where the full observation is incomplete—i.e., the partial point cloud may contain regions that are not present in the full reference point cloud (see Sec.~\ref{sec:results_incomplete}). Finally, ablation studies are conducted to demonstrate the effectiveness of each component in the proposed \emph{GAPR-Net}(see Sec.~\ref{sec:ablation}).

\subsection{Performance Comparison with Existing Methods}~\label{sec:results_sota}

%%%%%%%%%%%%%%%%%%%%%%%%%
\begin{figure*}[ht!]
\centering
\includegraphics[width=0.85\textwidth]{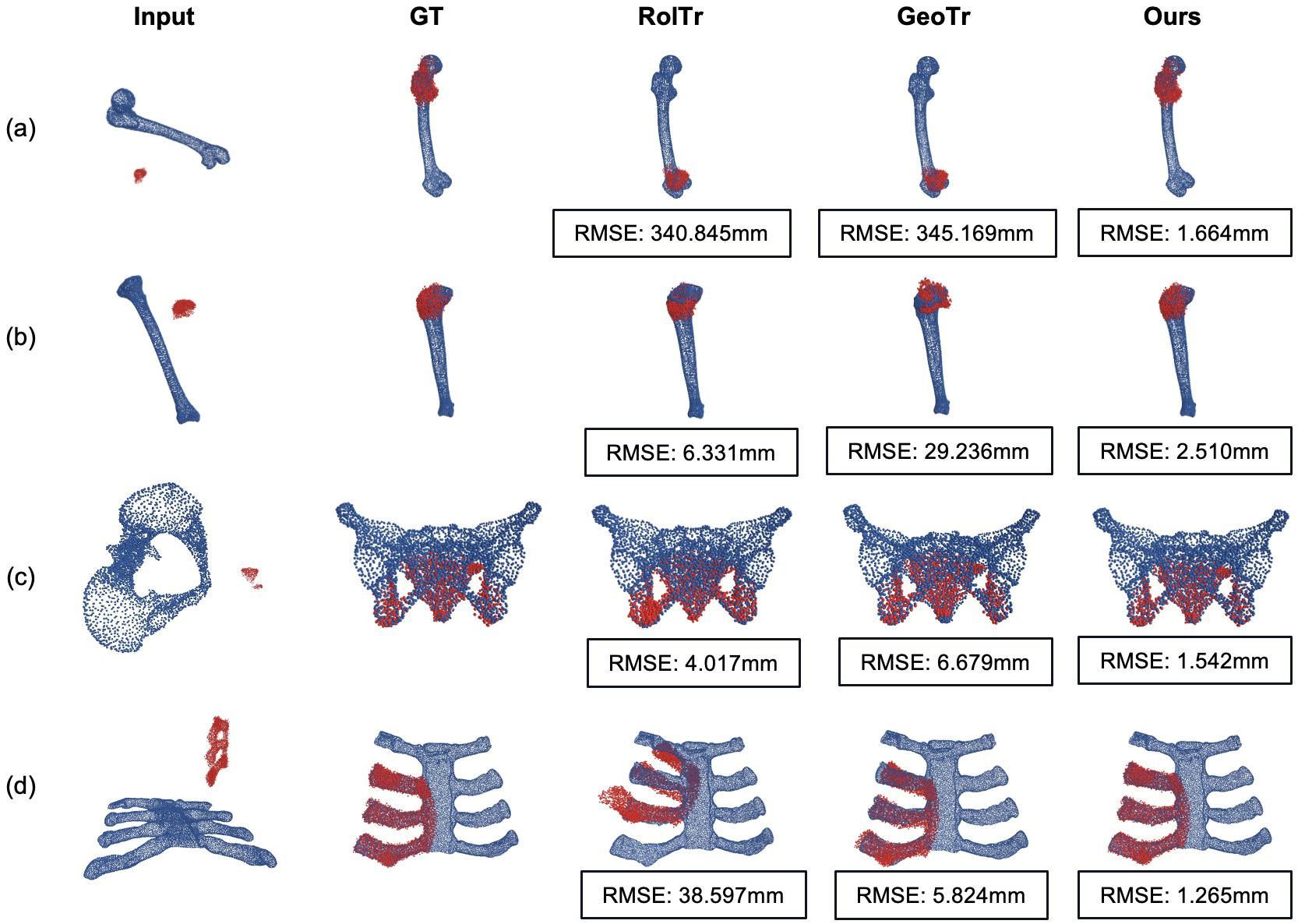}
\caption{Qualitative results of \emph{GAPR-Net} on partial-to-full registration for femur (a), tibia (b), pelvis (c), and thoracic cartilage (d), where the blue points represent the full point cloud and the red points represent the partial point cloud. Comparisons are conducted against GeoTransformer~\cite{qin2022geometric}, and RoITr~\cite{yu2023rotation}.
}
\label{fig_qualitive} 
\end{figure*}

\par
To evaluate the effectiveness of the proposed \emph{GAPR-Net} for partial-to-full point cloud registration, we compare it with the classical ICP algorithm~\cite{besl1992method} and some SOTA learning-based methods, including Predator~\cite{huang2021predator}, Lepard~\cite{li2022lepard}, GeoTransformer (GeoTr)~\cite{qin2022geometric}, and its extended variant RoITr~\cite{yu2023rotation}. In order to provide a fair comparison, we adopt four commonly used metrics according to~\cite{qin2022geometric, fu2021robust, wang2019prnet} in this study as well. First, RMSE refers to the average root mean square error between points in the original partial point cloud and the corresponding points after being transformed by the estimated transformation and then the inverse of the ground-truth transformation. Second, Chamfer Distance (CD) is defined as the sum of the average distances from each point in one cloud to its nearest neighbor in the other, computed in both directions. Only point pairs with distances less than 10.0 mm are considered. Thirdly, Registration Recall (RR) indicates the percentage of point cloud pairs whose normalized RMSE falls below a fixed threshold (RMSE~$<0.02$~mm after point cloud normalization). Lastly, the coefficients of determination for rotation R²($R$) and translation R²($t$) is computed from rotation angles (in degrees) over three axes, and R²($t$) from translations along three directions, where values closer to 1 indicate better alignment with the ground truth.

%%%%%%%%%%%%%%%%%%%%%%%%%%%

The average numerical results across all bone types—tibia, femur, pelvis, and thoracic cartilage—as well as the results for each individual type are summarized in Table~\ref{tab1_single_bone_comparsion}. In addition, a set of intuitive registration visualization results obtained by existing methods and the proposed method are demonstrated on four different types of bone in Fig.~\ref{fig_qualitive}. As shown in Table~\ref{tab1_single_bone_comparsion}, our method consistently outperforms all baselines across all registration metrics on the overall dataset. In particular, ICP and several learning-based methods, i.e., Predator~\cite{huang2021predator}, and Lepard~\cite{li2022lepard}, struggle to generalize to the partial-to-full registration setting, especially as reflected by their lower RMSE and RR values. In contrast, recent transformer-based approaches, GeoTransformer~\cite{qin2022geometric} and its variant RoITr~\cite{yu2023rotation}, demonstrate improved performance, achieving RR values of 90.7\% and 84.2\%, and RMSE of 2.337 mm and 3.103 mm, respectively. Benefiting from the integration of both global and local feature aggregation, our method achieves enhanced registration performance, which yields the best results in terms of RR, RMSE, R²($R$) and R²($t$)—94.2\%, 1.992 mm, 0.908 and 0.974, respectively. These results demonstrate that our method can outperform SOTA and result in more favorable registration accuracy.

\par
For registration on each individual bone type, our method achieves the best performance across all metrics on the pelvis and thoracic cartilage dataset, with particularly significant improvements for thoracic cartilage in RR—outperforming GeoTransformer by 7.3\% and RoITr by 42.5\%. These results highlight the effectiveness of our method in registering symmetric anatomical structures—such as the pelvis and thoracic cartilage—which remain a persistent challenge in point cloud registration. For the tibia and femur datasets, our method consistently achieves the best performance on critical registration metrics, including RR, R²($R$) and R²($t$), and maintains a comparable RMSE and chamfer distance. This demonstrates its effectiveness for different anatomies with varying geometries.

\newcommand{\pmc}[2]{#1\,$\pm$\,#2} % 定义简写命令，保持居中和间距

\begin{table*}[ht]
\centering
\renewcommand\footnoterule{\kern -1ex}
\resizebox{0.8\textwidth}{!}{
\begin{tabular}{c c c c c c c}
\toprule
Bone & Method & RR(\%) $\uparrow$ & CD(mm) $\downarrow$ & RMSE(mm) $\downarrow$ & $R^2$($R$) $\uparrow$ & $R^2$($t$) $\uparrow$ \\
\midrule
\multirow{6}{*}{Femur} &
ICP~\cite{besl1992method} & 8.4 & 9.115 & 157.942 & -0.367 & -19.021 \\
& Lepard~\cite{li2022lepard} & \pmc{7.6}{0.9} & \pmc{7.877}{0.012} & \pmc{40.146}{0.563} & \pmc{-0.009}{0.018} & \pmc{-8.051}{0.230} \\
& Predator~\cite{huang2021predator} & \pmc{78.8}{0.5} & \pmc{6.115}{0.016} & \pmc{3.510}{0.040} & \pmc{0.902}{0.031} & \pmc{0.950}{0.005} \\
& GeoTr~\cite{qin2022geometric} & \pmc{92.3}{0.9} & \pmc{5.762}{0.018} & \pmc{2.618}{0.073} & \pmc{0.882}{0.033} & \pmc{0.976}{0.003} \\
& RoITr~\cite{yu2023rotation} & \pmc{94.5}{0.4} & \textbf{\pmc{5.644}{0.011}} & \textbf{\pmc{2.252}{0.021}} & \pmc{0.887}{0.011} & \pmc{0.983}{0.001} \\
& Ours & \textbf{\pmc{95.6}{0.8}} & \pmc{5.679}{0.008} & \pmc{2.359}{0.046} & \textbf{\pmc{0.923}{0.033}} & \textbf{\pmc{0.984}{0.001}} \\
\midrule
\multirow{6}{*}{Tibia} &
ICP~\cite{besl1992method} & 4.2 & 8.232 & 137.278 & -0.445 & -19.813 \\
& Lepard~\cite{li2022lepard} & \pmc{7.1}{1.1} & \pmc{7.960}{0.042} & \pmc{40.364}{1.030} & \pmc{-0.028}{0.043} & \pmc{-9.065}{0.506} \\
& Predator~\cite{huang2021predator} & \pmc{74.8}{0.7} & \pmc{5.476}{0.013} & \pmc{5.574}{0.188} & \pmc{0.772}{0.036} & \pmc{0.535}{0.070} \\
& GeoTr~\cite{qin2022geometric} & \pmc{88.2}{2.4} & \pmc{5.099}{0.018} & \pmc{2.596}{0.089} & \pmc{0.883}{0.050} & \pmc{0.959}{0.006} \\
& RoITr~\cite{yu2023rotation} & \pmc{90.9}{0.2} & \textbf{\pmc{5.024}{0.004}} & \pmc{2.336}{0.009} & \pmc{0.871}{0.015} & \pmc{0.966}{0.004} \\
& Ours & \textbf{\pmc{93.6}{0.5}} & \pmc{5.032}{0.007} & \textbf{\pmc{2.331}{0.045}} & \textbf{\pmc{0.892}{0.016}} & \textbf{\pmc{0.971}{0.005}} \\
\midrule
\multirow{6}{*}{Pelvis} &
ICP~\cite{besl1992method} & 1.5 & 10.432 & 110.058 & -0.396 & -41.437 \\
& Lepard~\cite{li2022lepard} & \pmc{7.2}{0.5} & \pmc{7.961}{0.032} & \pmc{42.657}{1.274} & \pmc{0.013}{0.067} & \pmc{-9.111}{1.291} \\
& Predator~\cite{huang2021predator} & \pmc{58.9}{0.9} & \pmc{6.665}{0.012} & \pmc{12.963}{1.500} & \pmc{0.626}{0.050} & \pmc{-1.491}{0.563} \\
& GeoTr~\cite{qin2022geometric} & \pmc{99.1}{0.1} & \pmc{5.524}{0.011} & \pmc{1.493}{0.027} & \pmc{0.935}{0.033} & \pmc{0.991}{0.001} \\
& RoITr~\cite{yu2023rotation} & \pmc{84.8}{0.5} & \pmc{5.906}{0.013} & \pmc{2.802}{0.555} & \pmc{0.908}{0.054} & \pmc{0.979}{0.005} \\
& Ours & \textbf{\pmc{99.6}{0.1}} & \textbf{\pmc{5.430}{0.021}} & \textbf{\pmc{1.293}{0.021}} & \textbf{\pmc{0.942}{0.017}} & \textbf{\pmc{0.994}{0.001}} \\
\midrule
\multirow{6}{*}{Thoracic} &
ICP~\cite{besl1992method} & 0.5 & 6.010 & 50.838 & -0.278 & -10.015 \\
& Lepard~\cite{li2022lepard} & \pmc{7.0}{0.4} & \pmc{7.933}{0.056} & \pmc{42.054}{1.118} & \pmc{-0.005}{0.011} & \pmc{-9.027}{0.587} \\
& Predator~\cite{huang2021predator} & \pmc{30.3}{1.1} & \pmc{3.894}{0.008} & \pmc{13.497}{0.303} & \pmc{0.003}{0.043} & \pmc{-0.932}{0.054} \\
& GeoTr~\cite{qin2022geometric} & \pmc{82.3}{1.5} & \pmc{2.890}{0.009} & \pmc{2.550}{0.033} & \pmc{0.878}{0.013} & \pmc{0.883}{0.011} \\
& RoITr~\cite{yu2023rotation} & \pmc{47.1}{2.5} & \pmc{3.847}{0.055} & \pmc{7.055}{0.062} & \pmc{0.681}{0.027} & \pmc{0.751}{0.003} \\
& Ours & \textbf{\pmc{89.6}{1.4}} & \textbf{\pmc{2.803}{0.014}} & \textbf{\pmc{1.124}{0.041}} & \textbf{\pmc{0.962}{0.031}} & \textbf{\pmc{0.997}{0.001}} \\
\midrule
\multirow{6}{*}{Overall} &
ICP~\cite{besl1992method} & 2.9 & 8.515 & 126.460 & -0.345 & -20.340 \\
& Lepard~\cite{li2022lepard} & \pmc{7.5}{0.1} & \pmc{7.953}{0.018} & \pmc{39.839}{1.294} & \pmc{0.003}{0.025} & \pmc{-8.894}{0.381} \\
& Predator~\cite{huang2021predator} & \pmc{66.9}{1.0} & \pmc{5.571}{0.039} & \pmc{7.618}{0.553} & \pmc{0.660}{0.013} & \pmc{-0.178}{0.241} \\
& GeoTr~\cite{qin2022geometric} & \pmc{90.7}{0.4} & \pmc{5.001}{0.008} & \pmc{2.337}{0.021} & \pmc{0.884}{0.008} & \pmc{0.955}{0.002} \\
& RoITr~\cite{yu2023rotation} & \pmc{84.2}{0.2} & \pmc{5.171}{0.007} & \pmc{3.103}{0.056} & \pmc{0.866}{0.008} & \pmc{0.935}{0.002} \\
& Ours & \textbf{\pmc{94.2}{0.0}} & \textbf{\pmc{4.937}{0.007}} & \textbf{\pmc{1.992}{0.035}} & \textbf{\pmc{0.908}{0.010}} & \textbf{\pmc{0.974}{0.010}} \\
\bottomrule
\end{tabular}
}
\caption{Performance with Existing Methods. The standard deviation for ICP is excluded due to its deterministic nature.
}
\label{tab1_single_bone_comparsion}
\end{table*}

%%%%%%%%%%%%%%%%%%%%%%%%%%%%%%%%%%%%%%%%%
\subsection{Effect of Different Overlap Scenarios on Registration Performance}
% Effectiveness in Different Overlap Ratios}~\label{sec:results_overlap}
\par
This section further investigates the robustness of the proposed \emph{GAPR-Net} under varying overlap ratios, and a more challenging scenario where parts of the partial observation are not contained within the full point cloud.

%%%%%%%%%%%%%%%%%%%%%%%%%%%%%
\subsubsection{Performance Under Varying Partial Overlap Ratios}~\label{sec:results_overlap}
\par
In surgical settings, the proportion of the partial point cloud relative to the full point cloud typically varies. Generally, a smaller observed region leads to increased difficulty in achieving accurate registration. To assess the robustness of \emph{GAPR-Net} under varying degrees of partial observation, we evaluate its performance across different overlap ratios in the partial-to-full registration scenario. Specifically, the evaluation begins with an overlap ratio of 40\% and increases in increments of 10\% up to 70\%. The results, summarized in Table~\ref{tab3_overlap-ratio}. The results show that our method consistently achieves excellent registration performance for femur and pelvis across all overlap ratios, with RR exceeding 90\%. For tibia and thoracic cartilage, although performance slightly drops at lower overlap ratios (40\% and 50\%), it remains robust (RR $>$ 84\%). At higher overlap ratios (60\% and 70\%), RR rises to approximately 99\% and 94\%, respectively. These results demonstrate that \emph{GAPR-Net} maintains robust registration performance across a wide range of overlap conditions and is capable of achieving accurate alignment even in challenging cases with limited partial observations.

\begin{table}[ht]
\begin{center}
\renewcommand\footnoterule{\kern -1ex}
\renewcommand{\arraystretch}{1.3}
\resizebox{0.48\textwidth}{!}{
\begin{tabular}{c c c c c c}
\toprule 
Bone & Overlap & RR(\%) $\uparrow$ & RMSE(mm) $\downarrow$ & $R^2$($R$) $\uparrow$ & $R^2$($t$) $\uparrow$ \\
\midrule 
\multirow{4}{*}{Femur} 
& 40\% & \makecell{92.2} & \makecell{2.582} & \makecell{0.810} & \makecell{0.958} \\
& 50\% & \makecell{95.6} & \makecell{2.424} & \makecell{0.949} & \makecell{0.981} \\ 
& 60\% & \makecell{96.3} & \makecell{2.333} & \makecell{0.832} & \makecell{0.986} \\ 
& 70\% & \makecell{94.3} & \makecell{2.594} & \makecell{0.917} & \makecell{0.984} \\ 
\midrule
\multirow{4}{*}{Tibia} 
& 40\% & \makecell{85.8} & \makecell{2.703} & \makecell{0.821} & \makecell{0.956} \\
& 50\% & \makecell{93.3} & \makecell{2.329} & \makecell{0.854} & \makecell{0.974} \\ 
& 60\% & \makecell{91.7} & \makecell{2.351} & \makecell{0.895} & \makecell{0.979} \\ 
& 70\% & \makecell{96.0} & \makecell{2.123} & \makecell{0.903} & \makecell{0.982} \\ 
\midrule
\multirow{4}{*}{Pelvis} 
& 40\% & \makecell{98.3} & \makecell{1.459} & \makecell{0.905} & \makecell{0.986} \\
& 50\% & \makecell{99.1} & \makecell{1.412} & \makecell{1.000} & \makecell{0.993} \\ 
& 60\% & \makecell{99.4} & \makecell{1.295} & \makecell{0.991} & \makecell{0.995} \\ 
& 70\% & \makecell{99.9} & \makecell{1.203} & \makecell{0.913} & \makecell{0.995} \\ 
\midrule
\multirow{4}{*}{Thoracic} 
& 40\% & \makecell{85.4} & \makecell{1.320} & \makecell{0.802} & \makecell{0.996} \\
& 50\% & \makecell{84.0} & \makecell{1.514} & \makecell{0.999} & \makecell{0.994} \\ 
& 60\% & \makecell{94.7} & \makecell{0.941} & \makecell{1.000} & \makecell{0.996} \\ 
& 70\% & \makecell{93.2} & \makecell{1.078} & \makecell{0.860} & \makecell{0.997} \\ 
\bottomrule
\end{tabular}
}
\caption{Performance Under Varying Partial Overlap Ratios}
\label{tab3_overlap-ratio}
\end{center}
\end{table}

%%%%%%%%%%%%%%%%%%%%%%%%%
\begin{figure}[ht!]
\centering
\includegraphics[width=0.45\textwidth]{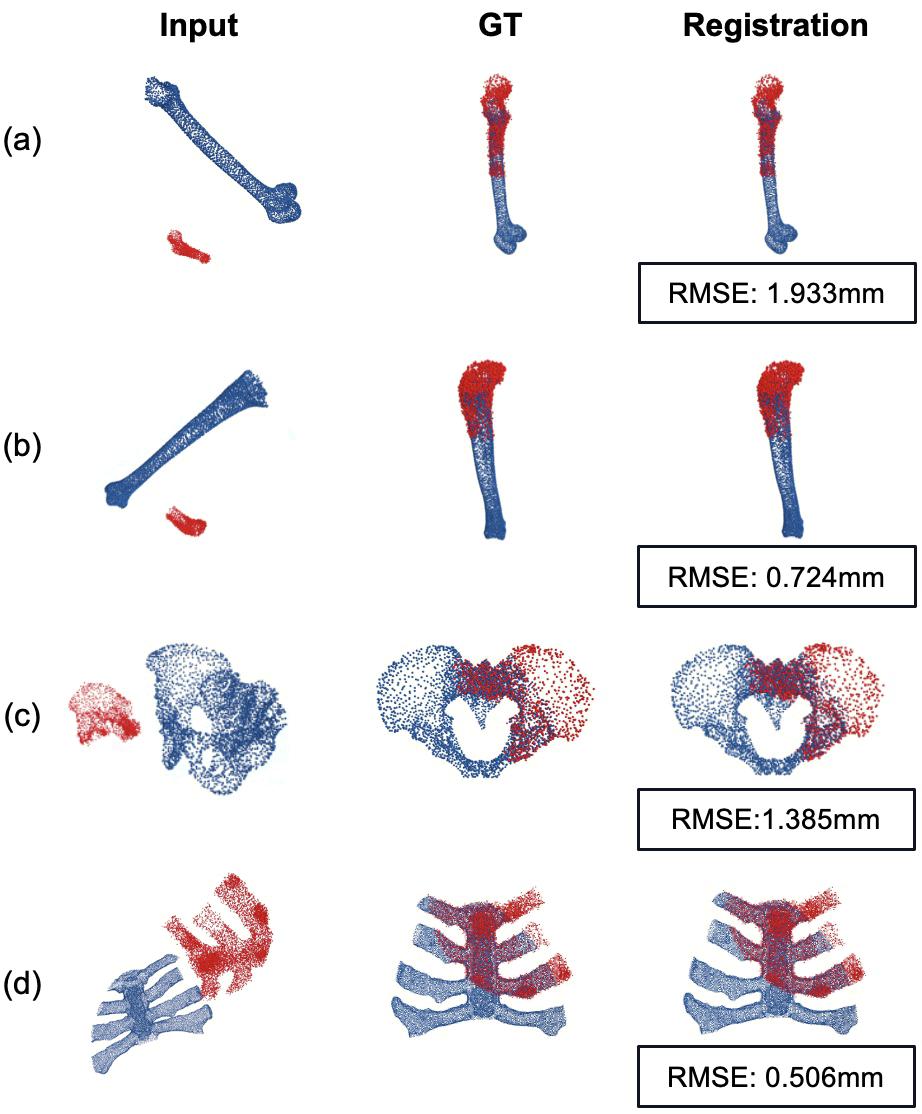}
\caption{Selected qualitative results of \emph{GAPR-Net} on partial-to-incomplete-full registration for the thoracic cartilage (a), pelvis (b), femur (c), and tibia (d), where the blue points represent the full point cloud and the red points represent the partial point cloud.}
\label{Fig_partial2partial}
\end{figure}
%%%%%%%%%%%%%%%%%%%%%%%%%

%%%%%%%%%%%%%%%%%%%%%%%%%
\subsubsection{Performance with Incomplete Full Point Cloud}~\label{sec:results_incomplete}
% \subsection{Effectiveness on Partial-to-Incomplete-Full Regitstation}~\label{sec:results_incomplete}
\par
Considering the practical challenges that partial observation may not always be entirely contained within the corresponding full point cloud or the preoperative point cloud, this section further investigate it. To simulate this challenging condition, we construct a partial-to-incomplete-full point cloud dataset following the pipeline outlined in Sec.~\ref{data-preprocessing}. Differently, to generate each pair, we randomly select two farthest points from the original point cloud using FPS~\cite{qi2017pointnet++} and apply KNN centered at each to extract the partial and incomplete full point clouds with retention ratios of 50\% and 90\%, respectively. This configuration ensures a minimum overlap of 40\% between the two point clouds and increases the likelihood that the partial point cloud is not entirely contained within the incomplete full point cloud, thereby creating a more realistic and challenging registration scenario.
\par
The quantitative results for individual bone types under the partial-to-incomplete-full registration scenario are summarized in Table~\ref{tab_partial-to-incomplete-full}, with representative qualitative examples shown in Fig.~\ref{Fig_partial2partial}. As reported in Table~\ref{tab_partial-to-incomplete-full}, the best performance is observed for the pelvis, achieving a registration recall (RR) of 98.1\%. For the tibia and femur, the RR values are comparable, around 90\%. The reduced performance relative to the pelvis is primarily attributed to their rounded geometry, which lacks distinctive features for robust matching. Notably, the thoracic cartilage exhibits slightly lower yet acceptable performance, with an RR of 86.2\%, which may be attributed to the relatively limited amount of training data. Overall, we consider the proposed \emph{GAPR-Net} demonstrates strong robustness in the partial-to-incomplete-full registration setting, achieving an overall RR of 90.7\% across all bone types, which is comparable to the averaged results 94.2\% in Table~\ref{tab1_single_bone_comparsion} when the partial point cloud is entirely contained in the full observation.

\begin{table}[ht]
\begin{center}
\renewcommand\footnoterule{\kern -1ex}
\renewcommand{\arraystretch}{1.3}
\resizebox{0.48\textwidth}{!}{
  \begin{tabular}{c c c c c c}
    \toprule 
    Bone Type & RR(\%) $\uparrow$ & CD(mm) $\downarrow$ & RMSE(mm) $\downarrow$ & $R^2$($R$) $\uparrow$ & $R^2$($t$) $\uparrow$ \\
    \midrule 
    Femur    & \makecell{89.7} & \makecell{6.207} & \makecell{2.739} & \makecell{0.945} & \makecell{0.969} \\
    Tibia    & \makecell{91.1} & \makecell{5.359} & \makecell{2.397} & \makecell{0.801} & \makecell{0.955} \\
    Pelvis   & \makecell{98.1} & \makecell{5.509} & \makecell{1.317} & \makecell{0.947} & \makecell{0.992} \\
    Thoracic & \makecell{86.2} & \makecell{3.318} & \makecell{1.562} & \makecell{0.856} & \makecell{0.993} \\
    Overall  & \makecell{90.7} & \makecell{5.271} & \makecell{2.206} & \makecell{0.923} & \makecell{0.976} \\
    \bottomrule
  \end{tabular}
}
\caption{Performance with Incomplete Full Point Cloud}
\label{tab_partial-to-incomplete-full}
\end{center}
\end{table}

\subsection{Ablation Study}
~\label{sec:ablation}
\par
We use a novel \emph{PGA}-based transformer in the KPConv backbone and the global aggregation, by which we achieve the state of the art. In the ablation study, we first explore the effectiveness of the backbone architecture combining the KPConv \cite{thomas2019kpconv} and \emph{PGA} by comparing our method with complete KPConv backbone and complete \emph{PGA} backbone. The result is shown in the upper part of Table~\ref{tab4_ablation_backbone}. Evidently, neither the standalone KPConv nor the \emph{PGA} backbone can match the performance achieved by their combination, with a gap of approximately 2\% in RR. This performance gain may be attributed to the complementary strengths of KPConv and Transformers, where their integration enhances local feature learning, ultimately improving the quality of registration.

\par
We further evaluate the effectiveness of the global \emph{cross-PGA} mechanism in the final cross-attention module. Specifically, we compare three variants: standard attention (StdA), \emph{PGA} without repositioning (\emph{PGA} w/o Rp), and \emph{PGA} with repositioning (\emph{PGA} w/ Rp). The results are summarized in the lower part of Table~\ref{tab4_ablation_backbone}. The results show that using standard attention in the final cross-attention module or introducing PGA without repositioning fails to achieve optimal registration performance. In contrast, incorporating PGA with repositioning consistently yields the best results across all evaluation metrics. These results suggest that after roughly aligning the point clouds, the relative positional embedding effectively captures meaningful relative geometry, thereby enhancing both matching quality and registration accuracy.

% \begin{table}[ht]
% \centering
% \caption{Ablation Study on Network Architecture}
% \renewcommand\footnoterule{\kern -1ex}
% \renewcommand{\arraystretch}{1.3}
% \resizebox{0.48\textwidth}{!}{%
%   \begin{tabular}{l*{6}{S}}
%     \toprule
%     {Backbone} & {RR(\%)} $\uparrow$ & {CD(mm)} $\downarrow$ & {RMSE(mm)} $\downarrow$ & {R\textsuperscript{2}($R$)} $\uparrow$ & {R\textsuperscript{2}($t$)} $\uparrow$\\
%     \midrule
%     KPConv & 88.8 & 5.032 & 2.241 & 0.878 & 0.973 \\
%     \emph{PGA} & 91.9 & 4.991 & 2.482 & 0.828 & 0.925 \\
%     KPConv+\emph{PGA} (Ours) & \bfseries 94.2 & \bfseries 4.937 & \bfseries 1.992 & \bfseries 0.908 & \bfseries 0.974 \\
%     \midrule
%     Final Cross-Attention \\
%     \midrule
%     Standard Attention & 91.2 & 5.049 & 3.803 & 0.830 & 0.577 \\
%     \emph{PGA} w/o Rp & 90.0 & 5.063 & 3.820 & 0.834 & 0.741 \\
%     \emph{PGA} w/ Rp (Ours)  & \bfseries 94.2 & \bfseries 4.937 & \bfseries 1.992 & \bfseries 0.908 & \bfseries 0.974 \\
%     \bottomrule
%   \end{tabular}
% }
% \label{tab4_ablation_backbone}
% \end{table}

\begin{table}[ht]
\centering
\renewcommand\footnoterule{\kern -1ex}
\renewcommand{\arraystretch}{1.3}
\resizebox{0.48\textwidth}{!}{%
  \begin{tabular}{l c c c c c}
    \toprule
    {Backbone} & {RR(\%)} $\uparrow$ & {CD(mm)} $\downarrow$ & {RMSE(mm)} $\downarrow$ & {R\textsuperscript{2}($R$)} $\uparrow$ & {R\textsuperscript{2}($t$)} $\uparrow$ \\
    \midrule
    KPConv              & \makecell{88.8}          & \makecell{5.032}          & \makecell{2.241}          & \makecell{0.878}          & \makecell{0.973}          \\
    \emph{PGA}          & \makecell{91.9}          & \makecell{4.991}          & \makecell{2.482}          & \makecell{0.828}          & \makecell{0.925}          \\
    KPConv+\emph{PGA} (Ours) & \makecell{\bfseries 94.2} & \makecell{\bfseries 4.937} & \makecell{\bfseries 1.992} & \makecell{\bfseries 0.908} & \makecell{\bfseries 0.974} \\
    \midrule
    \multicolumn{6}{l}{Final Cross-Attention} \\
    \midrule
    Standard Attention  & \makecell{91.2}          & \makecell{5.049}          & \makecell{3.803}          & \makecell{0.830}          & \makecell{0.577}          \\
    \emph{PGA} w/o Rp    & \makecell{90.0}          & \makecell{5.063}          & \makecell{3.820}          & \makecell{0.834}          & \makecell{0.741}          \\
    \emph{PGA} w/ Rp (Ours) & \makecell{\bfseries 94.2} & \makecell{\bfseries 4.937} & \makecell{\bfseries 1.992} & \makecell{\bfseries 0.908} & \makecell{\bfseries 0.974} \\
    \bottomrule
  \end{tabular}
}
\caption{Ablation Study on Network Architecture}
\label{tab4_ablation_backbone}
\end{table}

\par
We also investigate the effectiveness of our proposed \emph{PGA}-based transformer through an ablation study by comparing different transformer variants used in both the backbone and global aggregation modules, including GCN~\cite{huang2021predator}, Lepard~\cite{li2022lepard}, GeoTransformer~\cite{qin2022geometric}, RoITr~\cite{yu2023rotation} and Point Transformer V3 (PTV3)~\cite{Wu_2024_CVPR}. The results are presented in Table~\ref{tab5_ablation_trans}. The results demonstrate that our \emph{PGA}-based transformer consistently outperforms all compared methods across all evaluated metrics. These findings indicate that, compared to previously proposed transformer designs for point clouds, our \emph{PGA}-based transformer is more effective at capturing geometric features by introducing the geometry-aware positional embedding. This facilitates the network in predicting more accurate and reliable point correspondences.

% \begin{table}[ht]
% \centering
% \caption{Ablation Study on Transformer Variants}
% \renewcommand\footnoterule{\kern -1ex}
% \renewcommand{\arraystretch}{1.3}
% \resizebox{0.48\textwidth}{!}{%
%   \begin{tabular}{l*{5}{S}}
%     \toprule
%     {Transformer} & {RR(\%)} $\uparrow$ & {CD(mm)} $\downarrow$ & {RMSE(mm)} $\downarrow$ & {R\textsuperscript{2}($R$)} $\uparrow$ & {R\textsuperscript{2}($t$)} $\uparrow$\\
%     \midrule
%     GCN\cite{huang2021predator} & 80.8 & 5.232 & 3.182 & 0.885 & 0.918 \\
%     Lepard\cite{li2022lepard} & 92.4 & 5.007 & 2.875 & 0.868 & 0.908 \\
%     GeoTr\cite{qin2022geometric} & 91.5 & 4.988 & 2.411 & 0.889 & 0.951 \\
%     RoITr\cite{yu2023rotation} & 92.9 & 4.957 & 2.230 & 0.870 & 0.945 \\
%     PTV3\cite{Wu_2024_CVPR} & 91.1 & 4.981 & 2.557 & 0.875 & 0.899\\
%     \emph{PGA} (Ours) & \bfseries 94.2 & \bfseries 4.937 & \bfseries 1.992 & \bfseries 0.908 & \bfseries 0.974 \\
%     \bottomrule
%   \end{tabular}
% }
% \label{tab5_ablation_trans}
% \end{table}

\begin{table}[ht]
\centering
\renewcommand\footnoterule{\kern -1ex}
\renewcommand{\arraystretch}{1.3}
\resizebox{0.48\textwidth}{!}{%
  \begin{tabular}{l c c c c c}
    \toprule
    {Transformer} & {RR(\%)} $\uparrow$ & {CD(mm)} $\downarrow$ & {RMSE(mm)} $\downarrow$ & {R\textsuperscript{2}($R$)} $\uparrow$ & {R\textsuperscript{2}($t$)} $\uparrow$ \\
    \midrule
    GCN~\cite{huang2021predator}        & 80.8 & 5.232 & 3.182 & 0.885 & 0.918 \\
    Lepard~\cite{li2022lepard}           & 92.4 & 5.007 & 2.875 & 0.868 & 0.908 \\
    GeoTr~\cite{qin2022geometric}        & 91.5 & 4.988 & 2.411 & 0.889 & 0.951 \\
    RoITr~\cite{yu2023rotation}          & 92.9 & 4.957 & 2.230 & 0.870 & 0.945 \\
    PTV3~\cite{Wu_2024_CVPR}             & 91.1 & 4.981 & 2.557 & 0.875 & 0.899 \\
    \emph{PGA} (Ours)                   & \bfseries 94.2 & \bfseries 4.937 & \bfseries 1.992 & \bfseries 0.908 & \bfseries 0.974 \\
    \bottomrule
  \end{tabular}
}
\caption{Ablation Study on Transformer Variants}
\label{tab5_ablation_trans}
\end{table}

%%%%%%%%%%%%%%%%%%%%%%%%%
\section{Conclusion}
In this paper, we present \emph{GAPR-Net}, a novel coarse-to-fine framework for partial-to-full 3D point cloud registration, specifically designed to address challenges such as varying overlap ratios, fluctuating point densities, and noise commonly encountered in computer-assisted surgical scenarios. Unlike prior works that predominantly use transformers for global feature aggregation, \emph{GAPR-Net} introduces a hybrid design that integrates local convolution with a point-wise geometry-aware attention (\emph{PGA}) transformer. The \emph{PGA} module, built upon our proposed point-wise geometric feature (\emph{PGF}), effectively captures fine-grained geometric structures and enhances correspondence estimation. 

% To improve generalization across diverse anatomical categories, we further introduce an adaptive learning strategy tailored for multi-source datasets, enabling robust representation learning with limited medical data. 

Extensive experiments on a multi-class bone dataset—including pelvis, femur, tibia, and thoracic cartilage with overlap ratios between 40\% and 70\%—demonstrate that \emph{GAPR-Net} achieves state-of-the-art performance, with an overall registration recall (RR) of 94.2\%, RMSE of 1.992 mm, Chamfer Distance (CD) of 4.937, and R² values of 0.908 and 0.974 for rotation and translation, respectively. For the femur, \emph{GAPR-Net} matches the RMSE of RoITr~\cite{yu2023rotation} at 2.359 mm while achieving the highest RR of 95.6\%. For the tibia, pelvis, and thoracic cartilage, \emph{GAPR-Net} attains the best RR and RMSE scores: 93.6\% and 2.331 mm for tibia, 99.6\% and 1.293 mm for pelvis, and 89.6\% and 1.124 mm for thoracic cartilage. We further evaluate \emph{GAPR-Net} under challenging conditions, including low-overlap cases and scenarios where the full point cloud is incomplete, and the results remain robust. Ablation studies also confirm the superiority of the proposed \emph{PGA} transformer over existing transformer architectures.

To mimic real clinical challenges such as occlusions, limited field of view, and sensor noise, we randomly alter both the shape and density of the generated partial point cloud observations. The Experimental results demonstrate that our method provides a technically robust solution for partial-to-full registration while laying a solid foundation for precise robot-assisted CAS and autonomous robotic ultrasound scanning\cite{raina2024coaching} for further development. 

\bibliographystyle{IEEEtran}
\bibliography{references}

\end{document}